\documentclass[lettersize,journal]{IEEEtran}
\usepackage{amsmath,amsfonts}
\usepackage{algorithmic}
\usepackage{algorithm}
\usepackage{array}
\usepackage[caption=false,font=normalsize,labelfont=sf,textfont=sf]{subfig}
\usepackage{textcomp}
\usepackage{stfloats}
\usepackage{url}
\usepackage{verbatim}
\usepackage{graphicx}
\usepackage{caption}
\usepackage{adjustbox}
\usepackage{multirow}
\usepackage{cite}
\usepackage{hyperref}
\newcolumntype{R}[2]{%
    >{\adjustbox{angle=#1,lap=\width-(#2)}\bgroup}%
    l%
    <{\egroup}%
}

\usepackage{diagbox}

\usepackage[bottom]{footmisc}

\usepackage{hyperref}

\usepackage{amsmath,amsfonts,amssymb}

\begin{document}

\title{Forest Inspection Dataset for Aerial Semantic Segmentation and Depth Estimation}

\author{Bianca-Cerasela-Zelia Blaga and Sergiu Nedevschi
       
\thanks{The authors are with the Department of Computer Science, Technical University of Cluj-Napoca, 400114 Cluj-Napoca, Romania (e-mail: zelia.blaga@cs.utcluj.ro; sergiu.nedevschi@cs.utcluj.ro).}}

\maketitle

\begin{abstract}
Humans use UAVs to monitor changes in forest environments since they are lightweight and provide a large variety of surveillance data. However, their information does not present enough details for understanding the scene which is needed to assess the degree of deforestation. Deep learning algorithms must be trained on large amounts of data to output accurate interpretations, but ground truth recordings of annotated forest imagery are not available. To solve this problem, we introduce a new large aerial dataset for forest inspection which contains both real-world and virtual recordings of natural environments, with densely annotated semantic segmentation labels and depth maps, taken in different illumination conditions, at various altitudes and recording angles. We test the performance of two multi-scale neural networks for solving the semantic segmentation task (HRNet and PointFlow network), studying the impact of the various acquisition conditions and the capabilities of transfer learning from virtual to real data. Our results showcase that the best results are obtained when the training is done on a dataset containing a large variety of scenarios, rather than separating the data into specific categories. We also develop a framework to assess the deforestation degree of an area. 
\end{abstract}


\section{Introduction}
Worldwide efforts are being made to protect forests, slow the rate of deforestation, and reduce the negative impacts of environmental degradation. Researchers have successfully deployed Unmanned Aerial Vehicles (UAVs) to survey and monitor natural environments in GPS-denied environments since they are small, autonomous, and fast. For example, in \cite{tian} the authors construct a voxel representation of a forest, which helps the drone find a suitable trajectory, free of collisions. For forest inventory, the method SLOAM (Semantic Lidar Odometry and Mapping) \cite{chen} aims at estimating tree parameters using semantic segmentation and 3D point clouds. The authors also propose a method for semantic labeling of point clouds using virtual reality. \cite{kukko} is a methodology based on Graph SLAM that consists of extracting horizontal slices of point clouds to map the environment, then detecting tree trunks by looking for arcs of points, associating detections of the same tree, creating a graph representation of the initial trajectory and tree trunk detections, and correcting the trajectory with georeferenced measurements.

The latest developments in the field of Artificial Intelligence have brought improvements in the accuracy of algorithms that solve the problems of object detection and tracking, visual localization and mapping, trajectory planning, navigation control, and autonomous navigation. However, it comes at the cost of large amounts of training data that need to contain specific sensor recordings in the domain of interest. Since manual annotation is time-consuming, researchers have turned toward video game engines that are able to emulate real world scenarios, from urban cities to natural habitats. With the help of simulators, robots can navigate in these virtual environments and gather annotated data in real time. Thus, researchers have collected real-world datasets with manually annotated ground truth data that map urban environments \cite{tugraz, uavid, ruralscapes}, and have also created simulators that reproduce natural environments \cite{gazebo, sim4cv, airsim, carla} to automatically gather labeled information in the form of synthetic datasets \cite{midair, valid}.

\begin{figure}[]
	\includegraphics[width=\columnwidth]{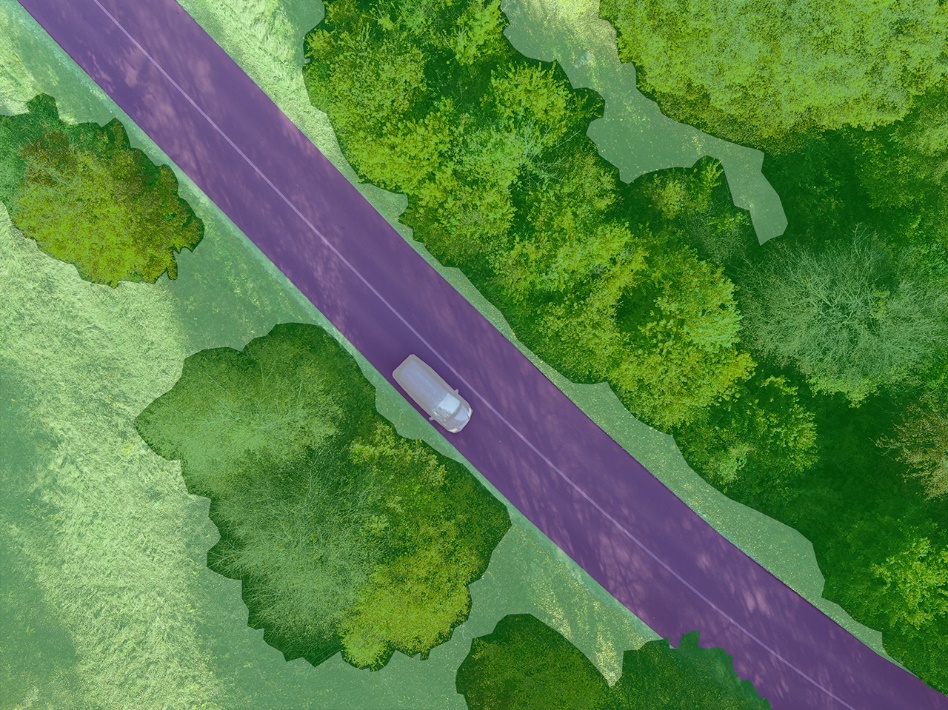}
	\centering
	\caption{Semantic segmentation labels overimposed on the camera recording from WildUAV.}
	\label{fig:image1}
	\vspace{-15pt}
\end{figure}

\begin{figure}[]
	\includegraphics[width=\columnwidth]{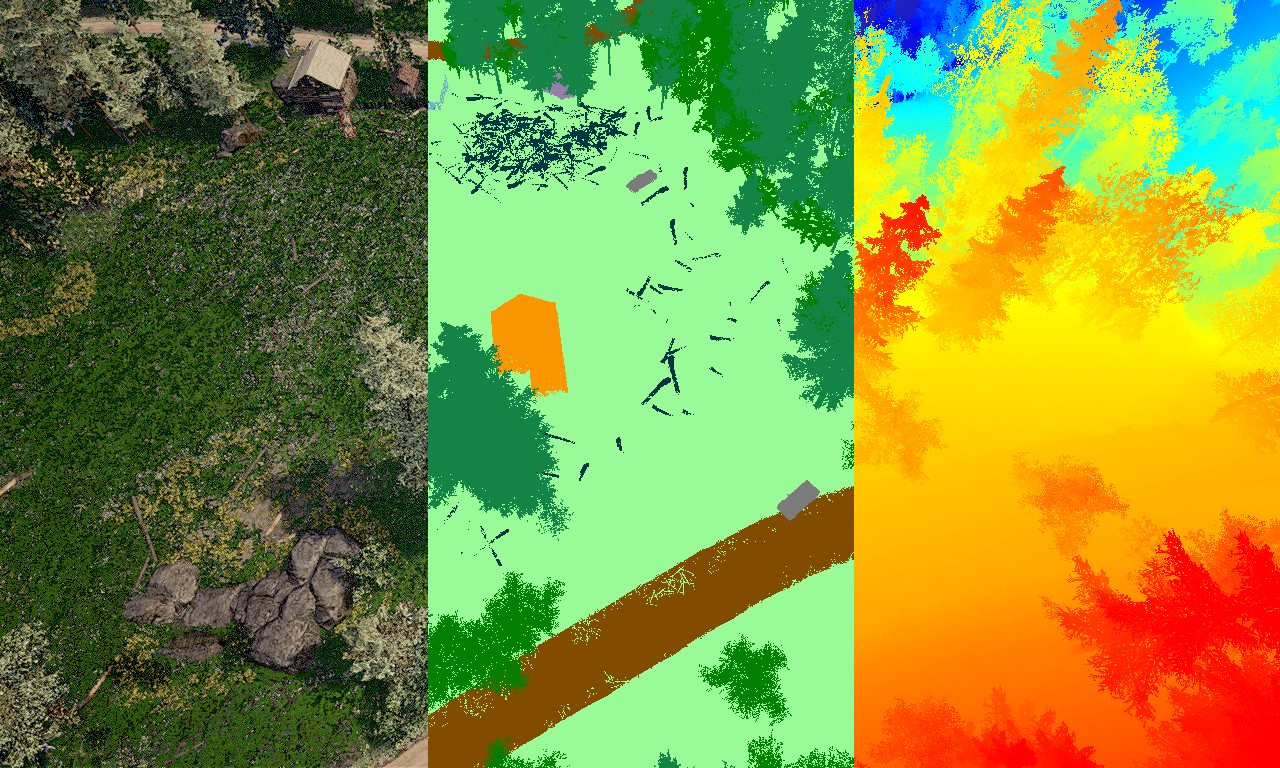}
	\centering
	\caption{Example of color, semantic labels, and depth map from the synthetic dataset.}
	\label{fig:image2}
	\vspace{-15pt}
\end{figure}

Our goal is to create an application focused on the visual inspection of forests, with the aim of monitoring deforestation. Existing datasets are limited in terms of sensor recordings and scenarios, therefore we created a new dataset, which contains real and virtual recordings. We worked on manually annotating real-world UAV recordings as seen in Figure \ref{fig:image1}, and we created a virtual forest environment using the AirSim simulator to record a large quantity of ground truth training data for semantic segmentation, depth estimation and autonomous navigation purposes, example which can be seen in Figure \ref{fig:image2}. We focused on creating a common ground between real and synthetic datasets, that complete each other and are semantically connected. Additionally, we paid attention to altitude, orientation degrees and illumination conditions, and carried out experiments to study the performance of semantic segmentation networks in various scenarios. Moreover, we worked on assessing the deforestation degree of an area of interest and developed a visual-based autonomous navigation algorithm. The contributions we bring in this paper are:

\begin{enumerate}
\item{We present the largest manually annotated semantic dataset for real-world recordings of natural environments. }
\item{We introduce a new synthetic dataset for forest monitoring, composed of color images, together with semantic labels, depth maps and camera positioning information, recorded at multiple altitudes, with camera orientation variation and in different lighting conditions.}
\item{We study the impact of various heights and recording angles for the task of semantic annotation.}
\item{We carry a performance analysis of multi-scale networks for aerial image segmentation.}
\item{We show the capabilities of transfer learning from synthetic to real data.}
\item{We propose a method for computing the deforestation degree of an area of interest.}
\end{enumerate}

We start by presenting, in Section \ref{rw}, state-of-the-art developments in the fields of drone datasets and multi-scale networks for semantic segmentation. Then in Section \ref{fi} we introduce the Forest Inspection dataset, detailing its contents and creation steps. Section \ref{me} is dedicated to the methodology, presenting framework components and explaining the experimental setup. The results are disseminated in Section \ref{re}, and in Section \ref{da} we discuss the deforestation methodology. We conclude the paper in Section \ref{conclusion}, discussing future work possibilities.

\section{Related Work} \label{rw}

\subsection{Drone Datasets for Forestry Applications}

The Semantic Drone Dataset provided by TUGRAZ \cite{tugraz} is focused on urban scenarios containing buildings, recorded from a bird’s eye view, at altitudes between 5 to 30 meters above the ground. There are 400 semantically annotated images, with a total of 20 classes, separated into categories such as building parts (roof, wall, window, door, fence), vehicles (car, bicycle), ground (dirt, gravel, rocks, paved area), vegetation (tree, grass), dogs and humans. The images are of high resolution, but the scenarios have little variation and are not suitable for our domain of interest. UAVid \cite{uavid} is a high-resolution UAV video dataset focused on the task of semantic segmentation of urban scenes. It contains 8 object categories and was recorded in 2 countries: China and Germany. Ruralscapes \cite{ruralscapes} is composed of over 800 manually annotated frames from an eastern European region. The authors also introduce a tool for semi-automatic video annotation which we have tested and found out that it performed poorly due to speed changes. Additionally, it lost smaller static objects such as people and equipment present in the scene.

The development of simulators has sped up the data gathering process by inserting robot controls in video game environments. Gazebo \cite{gazebo} is a robot simulation tool that provides a platform for testing algorithms, designing robots, and training agents in realistic environments. The simulator has a complex physics engine that supports robots like cars and quadrotors, together with a suite of sensors comprising monocular and stereo cameras, depth cameras, LiDARs, radars, sonars, GPS, and altimeters. In Gazebo, complex 3D environments can be created, in which the robots, objects and obstacles are placed. The biggest advantage that this simulator brings is the time it takes to create the desired setup for testing the implementation of algorithms. It works tightly with ROS, so deployment on real-life robots is fast. One disadvantage would be the low level of realism of urban scenarios. This depends on the objects imported into the project, and it is a time-consuming task to replicate the high level of complexity that cities in the real world display.

Sim4CV \cite{sim4cv} is a simulator built in Unreal Engine 4 that has a wide range of usability options for autonomous vehicles and UAVs. It is more suitable for reinforcement learning, and evaluation on pre-defined tracks. The available data provided is in the form of images, videos, image labels, segmentation classes, bounding box, depth, user input, physics and camera location. For UAVs, it is used to solve tasks like object tracking, 3D reconstruction, multi-agent collaboration, and aerial surveying. A map creation tool is provided, that includes elements like roads, trees, buildings etc. that can be used to create a new track. One disadvantage that we noticed is the slow speed of gathering data due to the high computational demands that use a lot of resources. Another disadvantage is the low level of realism and high level of repeatability that the provided environments bring.

CARLA \cite{carla} is a complex urban driving simulator, that uses Unreal Engine 4. It comes with 6 different towns, that combine modern and rural styles. The models for objects like buildings, cars, traffic signs or traffic furniture are diverse, making them look realistic. The physics engine and the Python API for control and recording tasks are easy to learn so the user can take recordings fast. CARLA also has a well-established semantic segmentation labeling sensor, that outputs annotated images which contain 13 classes inspired by the Cityscapes dataset \cite{cityscapes}. In our previous work \cite{blaga}, we introduced an aerial camera with which we recorded color images together with semantic labels, but we have found several inconsistencies and the movement of the camera does not replicate real-world drone physics. Another simulator built on top of Unreal Engine 4 is AirSim \cite{airsim}. This game engine comes with complex materials, lights, and shadows for photo-realistic imagery. It has a higher level of naturalness than the previously presented simulators. The drone physics are accurate and helpful to record aerial imagery, but the interactions are limited to gravity and simple collisions. It is the best simulator to consider since we can create our own virtual environment and set new semantic labels inside the video game engine, which we can use to record ground truth data for training deep learning algorithms.

Mid-Air \cite{midair} is a synthetic dataset for low-altitude drone flight, containing 54 sequences recorded using AirSim, in 4 weather conditions (sunny, cloudy, fog and sunset) and 3 seasons (spring, autumn and winter). It provides sequential training data containing color images, semantic segmentation labels, depth information and normal maps. The main drawback of this dataset is represented by the low altitude of the recordings, which resembles more the drone racing scenario than forest monitoring. Valid \cite{valid} is an aerial dataset for panoptic segmentation, that provides six different urban and natural virtual scenes. The ground truth data it provides comes from color images, together with bounding boxes, depth maps, and panoptic semantic segmentation labels. 

\subsection{Multi-Scale Networks for Semantic Segmentation}

One of the difficulties of solving the semantic segmentation task is the large variability of object sizes across the same image or even in a video sequence context. Therefore, researchers have worked on a way to fuse coarse, global information with fine, local information, as mentioned in \cite{zhao2019multi}. Traditional approaches rely on Maximum A Posteriori models, which means Bayesian networks of Markov Random Fields (MRF). Deep learning methods were combined with these two approaches to obtain better results. For example, in DeepLab \cite{chen2017deeplab}, MRF is applied as a post-processing step, while in \cite{zheng2015conditional}, MRF was used to pass the error information to the CNN. The authors of \cite{zhao2019multi} propose a multi-scale loss function that aims to accelerate a semantic segmentation network. They added intermediate mLoss layers to take advantage of multiple image scales, thus obtaining improved accuracy.

Dilated convolutions were one of the early methods to account for scale variations, as presented in \cite{yu2015multi}. These components have the advantage of capturing more information at the same cost as simple convolutions. Additionally, the receptive field is expanded, and more information is covered, which in turn leads to better image segmentation. The authors of \cite{audebert2016semantic} propose a method for segmenting earth observations taken from satellites by employing a framework composed of multi-kernel convolution layers for fastly aggregating multiple scales and data fusion of optical and laser recordings by residual correction. For the same purpose, in \cite{peng2019densely}, multi-scale filters are used to widen a dense connection and fully convolutional network (DFCN). 

Pyramid scene parsing has been applied with success to 3D structures in \cite{fang2019pyramid} by increasing the receptive field of points with contextual information, both regional and global. The results show a more accurate semantic segmentation of three-dimensional structures, improving large patches of object mislabeling. To enlarge the receptive field of a network, a two-stage multiscale architecture is proposed in \cite{ding2020semantic}. This brings considerable improvements over previous networks like FCN \cite{long2015fully}, U-Net \cite{ronneberger2015u}, PSPNet \cite{zhao2017pyramid}, SegNet \cite{badrinarayanan2017segnet} or DeepLabv3+ \cite{chen2018encoder} by being able to capture a larger receptive field and therefore more contextual information. The results highlight the increased segmentation accuracy this method brings, by being able to classify multiple regions of varying sizes. 

A new scheme entitled Multi-Scale Context Intertwining (MSCI) is presented in \cite{lin2018multi}. Bidirectional LSTM chains are used to propagate feature maps recurrently in an encoder-decoder network, and images are divided into super-pixels to use spatial relationship information to aggregate context information. In \cite{he2019dynamic}, the authors introduce the Dynamic Multi-scale Network (DMNet) that captures multi-scale content information for better pixel-level predictions. This framework is composed of parallel Dynamic Convolutional Modules (DCM), each estimating semantic representations at a specific scale. The output of these modules is further parsed by a final segmentation component, that combines the previously obtained feature vectors. The resulting segmentation maps have the advantage of being more consistent and less divided, with clearer borders. 

A method that is focused on boundary adherence is presented in \cite{pohlen2017full}. The methodology combines two computational streams, one that deals with the full image resolution to account for segment boundaries, and another one that employs pooling layers for robust feature recognition. These components are residually connected and the network has the advantage of computing features at multiple resolutions. DenseASPP (Atrous Spatial Pyramid Pooling) is presented in \cite{yang2018denseaspp}, which introduces dilation rates of multiple sizes at varying layers to capture and generate multiscale features, thus providing more dense information that is better classified. 

A breakthrough in this domain has been brought by High-Resolution Network (HRNet) \cite{wang2020deep}, which obtained top results in multiple fields: human pose estimation, facial landmark detection, image classification, semantic segmentation, and object detection. The network uses multiple convolutional layers that maintain a high-resolution representation throughout the network, ensuring that no data is lost, which often happens in networks that downsample the input image and then upsample the computed feature vectors to extract the semantic information. We consider this network for further study and perform experiments on it to test its performance for the task of segmenting images recorded from drones.

\subsection{Solutions for Aerial Semantic Segmentation}

Works in the area of semantic segmentation of images recorded from UAVs focus on attention-based modules \cite{mou2020relation,li2021multitask,deng2021ccanet,niu2021hybrid}, on extracting multi-scale features \cite{he2016deep,szegedy2017inception,chen2018encoder,kirillov2019panoptic} or highlighting boundary information \cite{kirillov2020pointrend,mi2020superpixel,zhen2020joint,sun2020bas,feng2020npaloss,li2021pointflow}. To solve the task of high-altitude aerial segmentation, the authors of \cite{uavid} have designed a Multi-Scale Dilation network, which combines Fully Convolutional Network, Dilation Network and U-Net architecture to improve the prediction at various scales. The network presented in \cite{liu2021high} is an encoder-decoder framework based on Adaptive Multi-scale Modules to segment objects of different sizes and an Adaptive Fuse Module that combines shallow and deep features. Two attention modules are employed, one for gathering spatial information, and one for modeling channel connections. PointFlow Network \cite{li2021pointflow} is based on a Feature Pyramid Network, with the novelty coming from a Point Flow Module that is inserted between consecutive layers of the decoder to improve the segmentation results by combining the boundary and salient regions.

\section{Forest Inspection Dataset } \label{fi}

We created a large aerial dataset combining manually annotated real-world recordings and synthetic data collected with the use of simulators to solve tasks necessary for forest monitoring, such as semantic segmentation, depth estimation and scene understanding. To our knowledge, it is the first dataset of this type that contains densely annotated semantic images which include the class of fallen trees, together with the depth and positioning information.

\subsection{Real Dataset}

Starting from the real-world recordings from WildUAV\footnote{The WildUAV dataset can be downloaded from the GitHub repository: \href{https://github.com/hrflr/wuav}{https://github.com/hrflr/wuav}} \cite{wilduav}, we have manually annotated 4 mapping sets using Adobe Photoshop CS6 \cite{pscs6}, resulting in dense and accurate semantic labels. To prevent the blending of colors, we worked with the Pencil Tool both for Brush and Eraser, and we deselected the anti-aliasing option for the lasso tools and the magic wand tool. 

\begin{figure}[ht]
	\includegraphics[width=\columnwidth]{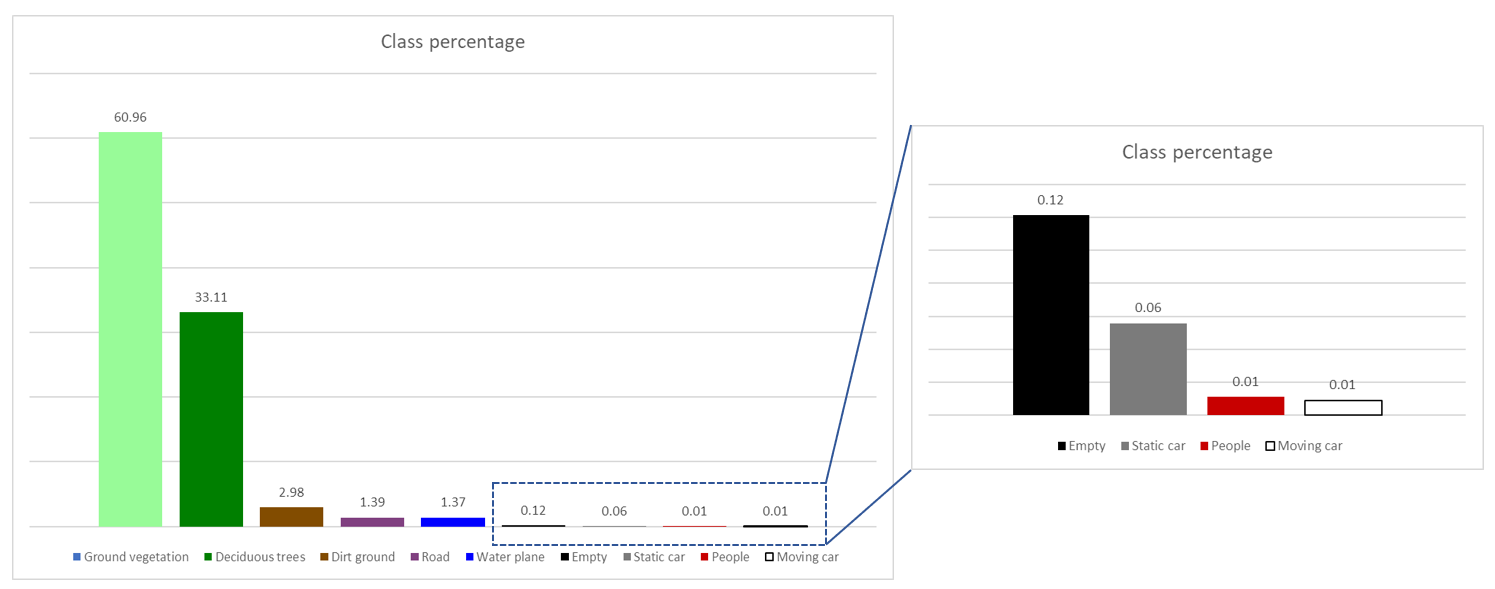}
	\centering
	\caption{Distribution of semantic classes in the WildUAV real dataset.}
	\label{fig:real_class}
\end{figure}

\begin{figure}[ht]
	\includegraphics[width=\columnwidth]{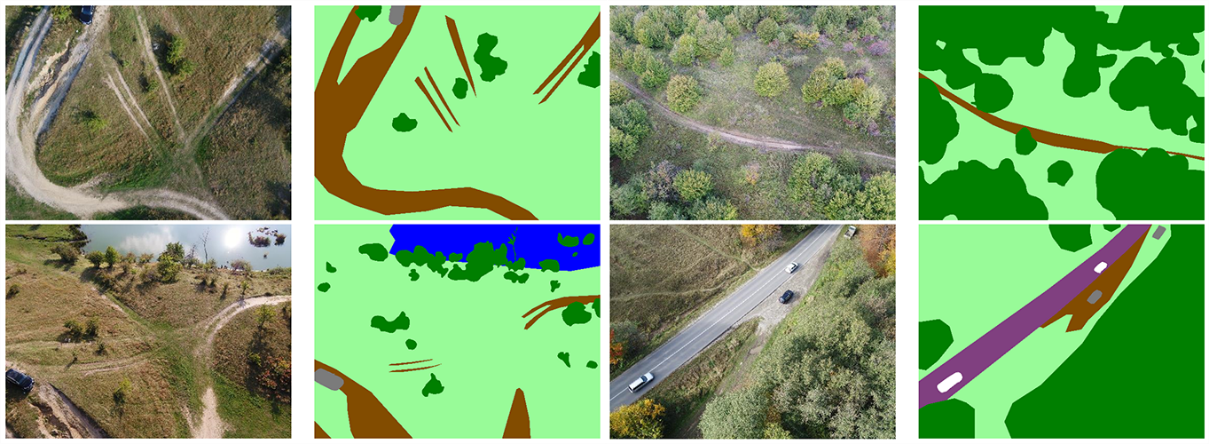}
	\centering
	\caption{Pairs of manually annotated images from WildUAV, showcasing the large variability of terrain and class labeling.}
	\label{fig:real_data}
\end{figure}

Each class is represented by a separate instance layer, and the objects were annotated either using a mouse with the polygonal lasso or with a Wacom Bamboo tablet with the free-hand lasso tool. We also used an automatic label propagation methodology for videos \cite{ruralscapes}. Thus we obtained over 2,600 real semantic segmentation images with a number of 6 classes present in the annotations, whose numerical distribution can be seen in Figure \ref{fig:real_class}. Some examples extracted from the dataset are shown in Figure \ref{fig:real_data}, offering a glimpse into the practical application and diversity of the annotated data.

\subsection{Synthetic Dataset}

Since data collection and annotation is time-consuming, we turned our attention to simulators, which can be used to record sets that meet specific criteria and provide access to APIs that enable drone control. We started with the creation of a virtual environment in Unreal Engine 4, using the European Forest package \cite{europeanforest} available on the Unreal Engine Marketplace, which provides scanned, high-quality foliage textures. There are 3 categories of trees: birch, fir and Scots pine, each with multiple variations of shape and size, along with fallen trunks and a wide variety of ground vegetation instances. We added several vehicles throughout the map, using the models provided in the Vehicle Pack \cite{vehiclepack}.

\begin{table}[!ht]
\centering
\caption{WildUAV dataset contents and number of annotations for the recordings.}\label{tab:results}
\begin{tabular}{ll}
\textbf{Dataset} & \textbf{\#images} \\
\hline
seq00   & 244      \\
seq01   & 497      \\
seq02   & 356      \\
seq03   & 442      \\
vid02   & 1,069    \\
\hline
\textbf{Total}   & \textbf{2,608}  
\end{tabular}
\end{table}

In the virtual environment, we selected the AirSim module as drone control and set a class ID for each object in the map, by modifying the \textit{CustomDepth Stencil Value} field. To change the colors of the labels we modified the asset named \textit{seg\_color\_pallete} to our desired values. To adjust the illumination conditions for the overcast scenario, we modified the Light Source, SkySphere, SkyAtmosphere and SkyLight. The drone mapping methodology is similar to real-world recordings.  The drone starts from the center of the map, moves to the first corner of the area of interest and makes several linear passes over the forest environment. The exact values we used were box size of 170, stripe width of 50, altitude values of 30, 50 and 80, in meters, and speed of 5 m/s (meters per second). 

\begin{figure}[t]
	\includegraphics[width=\columnwidth]{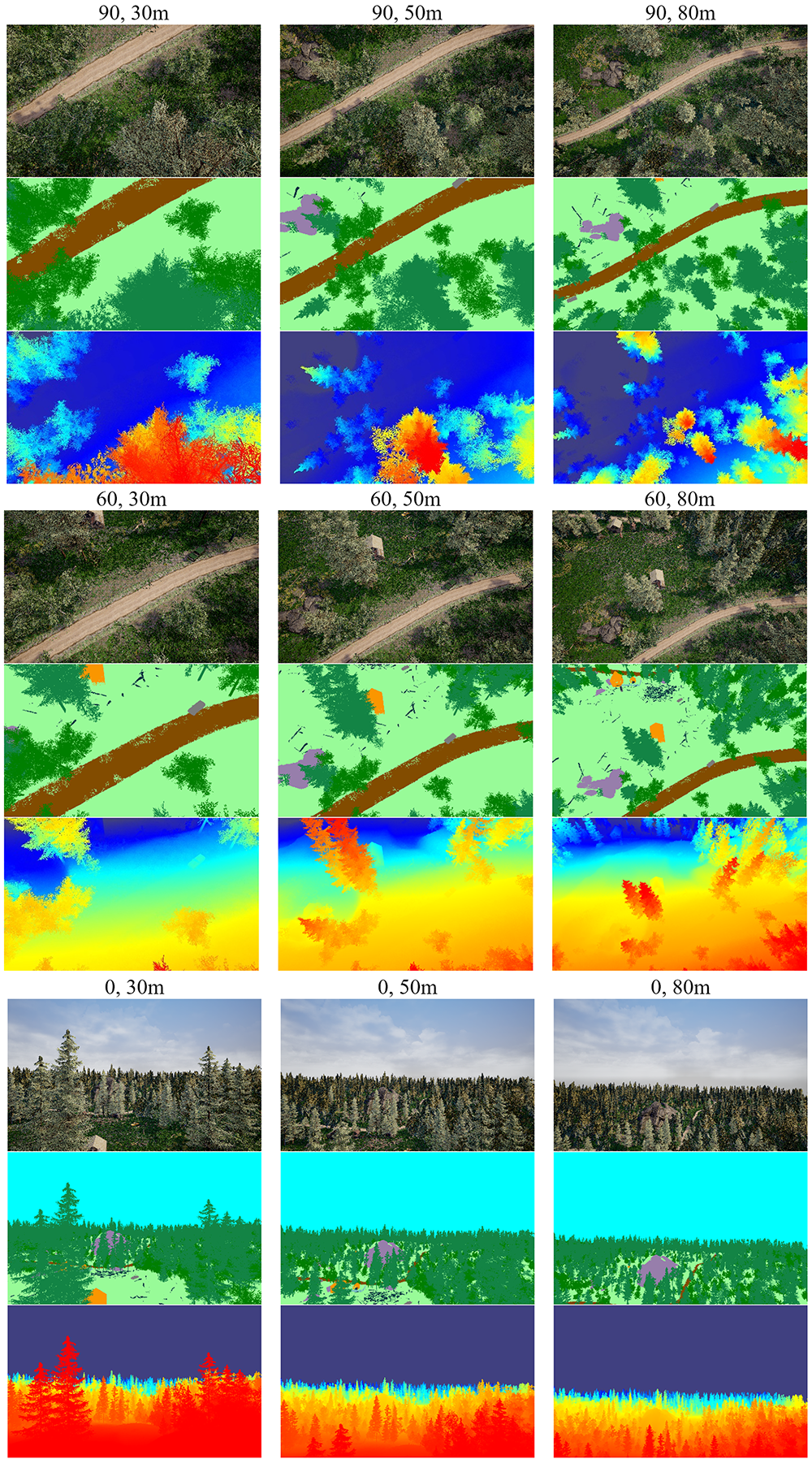}
	\centering
	\caption{Examples of images recorded by drone in the virtual environment, with angle and altitude variations, including color, semantic labels, and depth map information.}
	\label{fig:synth_data}
	\vspace*{-3mm}
\end{figure}

We recorded 22 sequences containing over 31,000 color images, along with semantic segmentation and depth information. A total of 11 classes are present as seen in Figure \ref{fig:synth_class}, with the caveat that we distinguish between conifers, deciduous and fallen trees. Additional drone positioning and orientation data are provided by the AirSim controller. The Forest Inspection dataset\footnote{The contents will be available for download from \href{https://www.dropbox.com/scl/fo/unxtplo0ge2qch022smbu/h?rlkey=ldllwy4zxqx0jasyj0v5ek7w8&dl=0}{Dropbox}.}  was recorded in 2 weather conditions: sunny and overcast, from altitudes of 30, 50 and 80 meters, with 3 degrees of pitch angle (0, 60, and 90), as shown in Figure \ref{fig:synth_data}.

\begin{table}[t]
\centering
\caption{Dataset contents and number of annotations for the synthetic  recordings in the Forest Inspection dataset.}\label{tab:results}
\begin{tabular}{llll}
Sunny FOV 90    &            &             &          \\
Sequence\#      & Pitch(deg) & Altitude(m) & \#images \\
\hline
seq1            & 0          & 30          & 1,468    \\
seq2            & -60        & 30          & 1,501    \\
seq3            & -90        & 30          & 1,469    \\
seq4            & 0          & 50          & 1,432    \\
seq5            & -60        & 50          & 1,447    \\
seq6            & -90        & 50          & 1,437    \\
seq7            & 0          & 80          & 1,436    \\
seq8            & -60        & 80          & 1,464    \\
seq9            & -90        & 80          & 1,397    \\
                &            &             &          \\
Overcast FOV 90 &            &             &          \\
Sequence\#      & Pitch(deg) & Altitude(m) & \#images \\
\hline
seq10           & 0          & 30          & 1,522    \\
seq11           & -60        & 30          & 1,618    \\
seq12           & -90        & 30          & 1,556    \\
seq13           & 0          & 50          & 1,398    \\
seq14           & -60        & 50          & 1,398    \\
seq15           & -90        & 50          & 1,341    \\
seq16           & 0          & 80          & 1,489    \\
seq17           & -60        & 80          & 1,471    \\
seq18           & -90        & 80          & 1,398    \\
                &            &             &          \\
Sunny FOV 60    &            &             &          \\
Sequence\#      & Pitch(deg) & Altitude(m) & \#images \\
\hline
seq19           & -90        & 30          & 1,325    \\
seq20           & -90        & 30          & 1,324    \\
seq21           & -90        & 30          & 1,322    \\
seq22           & -90        & 30          & 1,309    \\
                &            &             &          \\
     
                &            & \textbf{Total}       & \textbf{31,531}  
\end{tabular}
\end{table}

\begin{figure}[t]
	\includegraphics[width=\columnwidth]{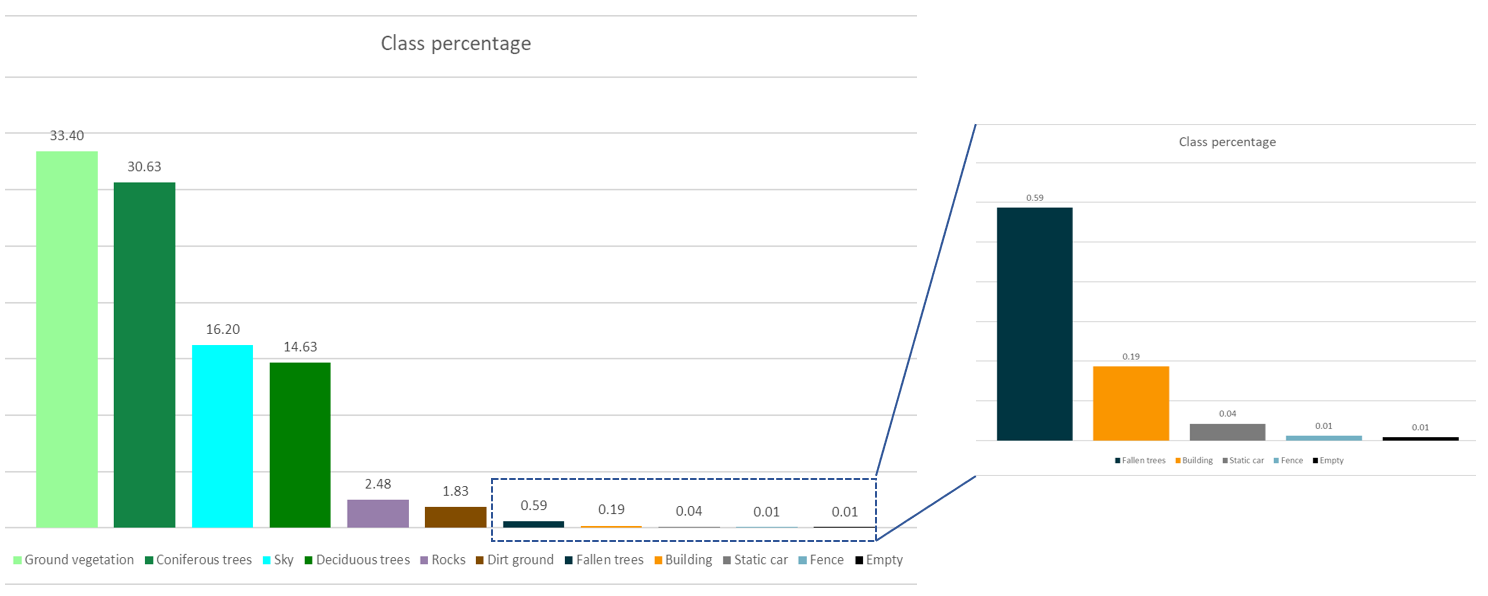}
	\centering
	\caption{Distribution of semantic classes in the Forest Inspection synthetic set.}
	\label{fig:synth_class}
\end{figure}

\section{Methodology for Semantic Segmentation of Aerial Images} \label{me}

\begin{table*}[t]
\begin{center}
\centering
\begin{tabular}{c}
\adjustimage{width=18cm, valign=m}{hrnet} \\
a) HRNet architecture \\ 
\adjustimage{width=14cm, valign=m}{pfnet} \\
b) PointFlow Network Architecture \\
\end{tabular}
\end{center}
\captionof{figure}{The architectures of the two networks. Both use a multi-scale framework, a) HRNet employs a sequential flow, while b) PointFlow Net uses a pyramidal structure.} \label{fig:sos}
\vspace*{-3mm}
\end{table*}

\subsection{HRNet}

The structure of HRNet \cite{wang2020deep} is composed of several stages (usually 4), where each n-th stage contains n streams corresponding to n resolutions. Multi-resolution fusions are repeatedly computed by exchanging feature maps across parallel streams. The main two advantages of this framework are the conservation of high-resolution representations throughout the network, and the learned representation is spatially more precise according to the authors of the method. 

For the task of semantic segmentation, HRNetV2p is used. Compared to state-of-the-art methodologies, this new architecture achieves better results with fewer parameters, lower computational complexity, and decreased inference time. An important factor to notice is the computation of high-resolution sub-network in parallel with lower resolution sub-network. The information from lower layers is fused at top level, where different strided convolutions of sizes $3\times3$ and $1\times1$ are applied from left to right to combine information first from medium and low layers to compute top information, from medium and low layers to output the middle feature vector, and lastly from all three layers to compute the low-level information. Thus, each output representation is the sum of three inputs. The softmax loss is used to predict the output segmentation.

\subsection{PointFlow Network}

PFNet \cite{li2021pointflow} has the architecture of Feature Pyramid Network (FPN), with the bottom-up part computing feature maps for lower-level semantics with more accuracy since the input is subsampled fewer times, the top-down part for computing higher resolution features by using PointFlow Modules. The encoder and decoder are connected by a Pyramid Pooling Module (PPM), which computes various feature maps at different average pooling sizes. The core part of PFNet is the PointFlow module which is composed of two modules: a Dual Point Matcher and a Dual Region Propagation. The first one computes the corresponding key points between two maps, while the latter computes salient and boundary point flows. The Dual Point Matcher has two parts, for generating the salient map and the sampled indexes using the Dual Index Generator. 

\textbf{Feature Pyramid Network.} Consists of an encoder-decoder structure for extracting features. The bottom-up part computes feature maps for lower-level semantics, but it is more accurate since the subsampling operation is done fewer times. As the spatial resolution decreases, the semantic value increases. The top-down flow is for computing higher resolution features by upsampling. The lateral connections used for merging feature maps of the same spatial size are replaced by a PointFlow Module which combines different sizes of feature maps. 

\textbf{Pyramid Pooling Module (PPM).} Takes as input a feature map and performs multiple average pooling operations ($1\times1$, $2\times2$, $3\times3$ and $6\times6$) to sample the space in various sub-region sizes.

\textbf{Dual Point Matcher.} Find corresponding points between two maps, one from the previous encoder path. and one from the current decoder flow. First, the salient map is computed by combining two feature maps. Then, by using the Dual Index Generator, sampled indexes are computed.

\textbf{Dual Index Generator.} Computes the boundary and salient points. For computing the boundary, Canny edge detection is applied on the label mask, while the salient map is obtained by subtracting the boundary from the feature map. 

\textbf{Sampling.} In PointFlow Network, rather than using dense affinity, the authors use a point sampler to select matched representative points to balance the foreground-background information. Each feature map is processed to extract low and high edge features, respectively low and high salient features. 

\textbf{Dual Region Propagation.} Computes the salient and boundary point flows based on the previously computed point flows. The final values are a combination of semantic feature maps, with improved boundary and salient values.

The binary BCE loss is used for edge prediction in the PFM module, while the cross-entropy loss is employed for the semantic segmentation prediction.

\begin{figure*}[!ht]
	\includegraphics[width=18cm]{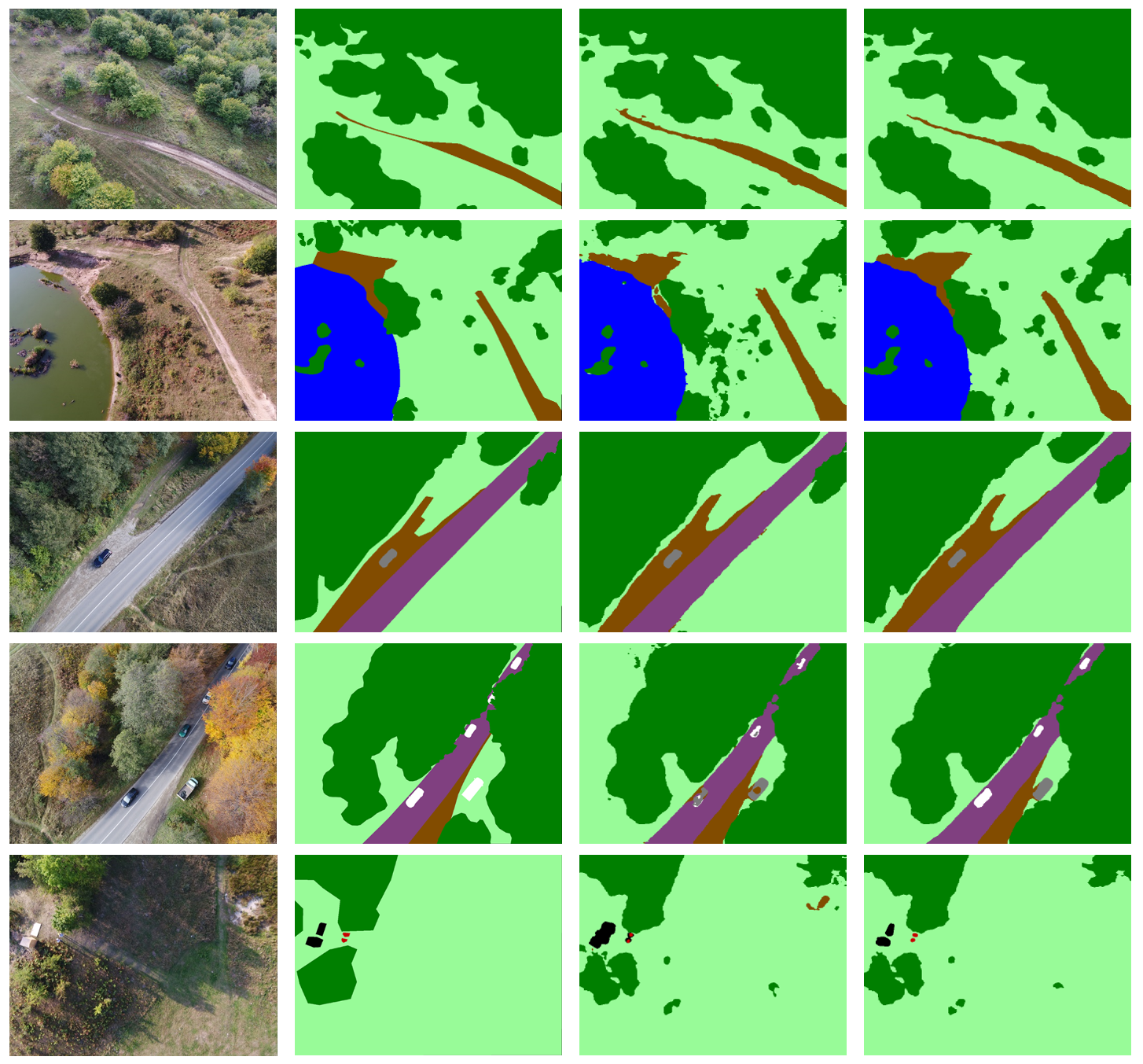}
	\centering
	\caption{Semantic segmentation results on WildUAV. From left to right: color image, ground truth labels, HRNet and PointFlow network results.} 
	\label{fig:synth_class}
\end{figure*}

\begin{table*}[!ht]
\centering
\caption{Semantic segmentation results for the WildUAV real drone recordings.}\label{tab:wilduav_results}
\begin{tabular}{c|l|ccccccccc|c}
& Class         & Grd. veg. & Trees & Dirt ground & Road  & Water & Empty & Static car & Moving car & People & Overall \\
\hline
\multirow{2}{1.5cm}{Semantic segmentation} & HRNet         & 93.40             & 89.64           & 79.23       & 92.69 & \textbf{95.18} & 62.61 & 81.58      & 28.45  &  27.63     & 72.27   \\
& PointFlow & \textbf{94.12}   & \textbf{91.52} & \textbf{82.92} & \textbf{93.01} & 94.78 & \textbf{64.93 }& \textbf{85.27 }     & \textbf{34.63}  & \textbf{30.85}      & \textbf{74.67} \\
\hline
\multirow{2}{1.5cm}{Transfer learning}   
& HRNet         & 94.00               & 89.94           & 80.13       & 92.99 & 96.38 & 63.21 & 82.78      & 29.05  & 27.93      & 72.93   \\
& PointFlow & \textbf{95.40}            &\textbf{91.74}         &\textbf{81.93}       & \textbf{95.99} & \textbf{97.58} & \textbf{66.21} & \textbf{85.18}      & \textbf{34.28}  & \textbf{31.53}   & \textbf{75.54} 
\end{tabular}
\end{table*}

\subsection{Training Scenarios}

To perform a comparison between HRNet and PointFlow network, we perform the following experiments:

\begin{enumerate}
\item{on real data - using only the annotation from WildUAV, more specifically mapping sequences seq00, seq01, seq02, and seq03;}
\item{on synthetic data - using the virtual recordings from the simulator, for the sequences from Sunny FOV 90, and Overcast FOV 90;}
\item{transfer learning - to observe the usefulness of the virtual data in a real-world scenario, we first train the networks on the synthetic sequences that have the camera oriented downward, for pitch angles of -60° and -90°, respectively with numbers 2, 3, 5, 6, 8 and 9, followed then by the real drone recordings;}
\item{ablation study - to asses the performance of each network in terms of number of parameters, model size and inference time; additionally, we study the impact of various recording conditions on the segmentation results. }
\end{enumerate}

\begin{figure*}[!ht]
	\includegraphics[width=18cm]{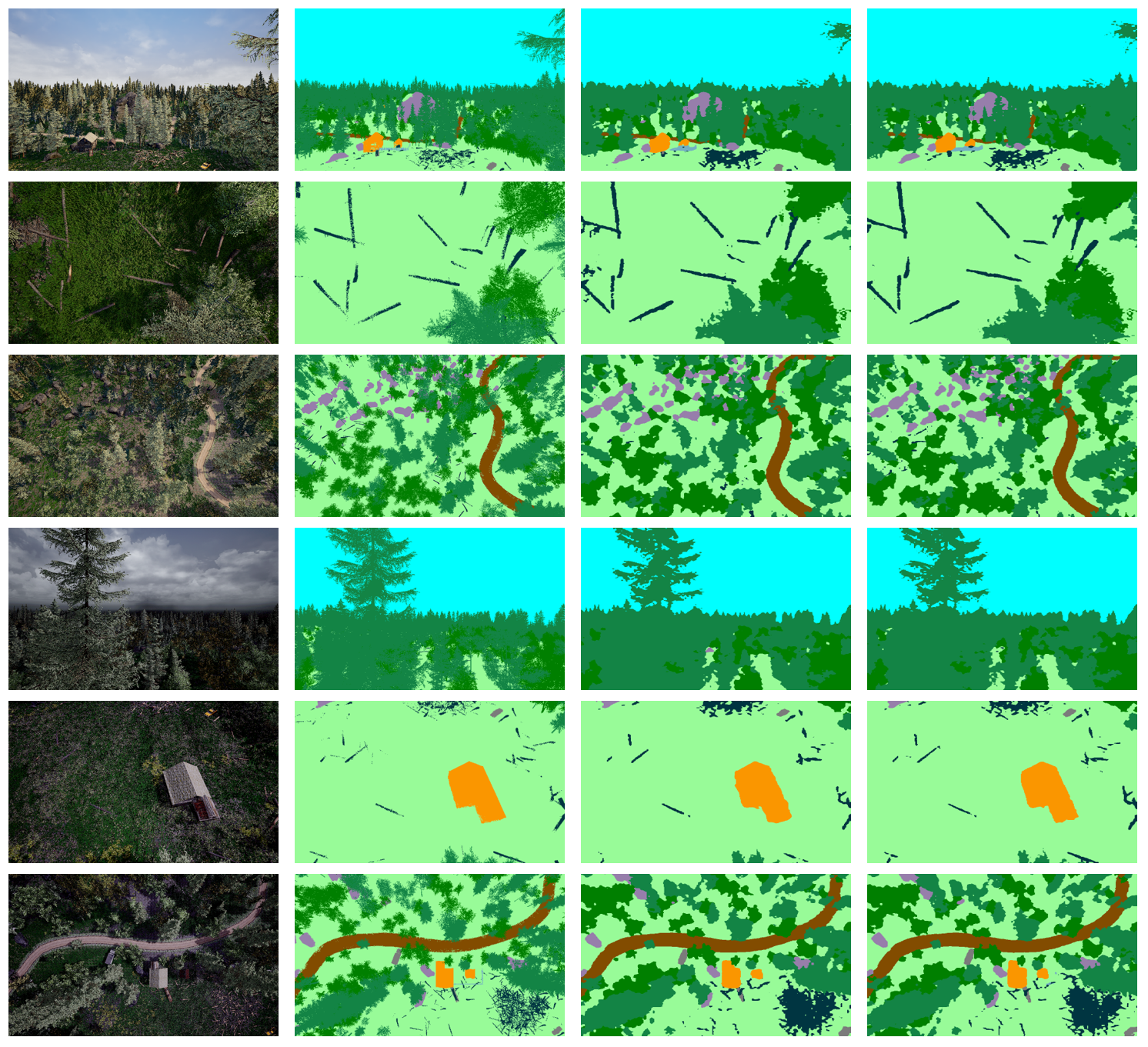}
	\centering
	\caption{Semantic segmentation results on Forest Inspection Dataset, on sunny and overcast lighting conditions. From left to right: color image, ground truth labels, HRNet and PointFlow network results.}
	\label{fig:fi_res}
\end{figure*}

\begin{table*}[!ht]
\centering
\caption{Semantic segmentation results for the Forest Inspection virtual aerial dataset, during sunny and overcast illumination conditions.}\label{tab:fi_sunny}
\begin{tabular}{c|l|cccccccccc|c}
& Class         & Grd. veg. & Deciduous  & Coniferous  & Fallen trees & Sky   & Rocks & Dirt ground & Building & Fence & Car & Overall \\
\hline
\multirow{2}{0.8cm}{Sunny} & HRNet         & 81.16             & 68.54           & 78.53            & 44.40        & 97.37 & 76.61 & 81.83       & 82.56    & 38.89 & 58.48      & 70.84   \\
& PointFlow & \textbf{83.94}            & \textbf{69.03}           & \textbf{79.41}           &\textbf{47.56}        & \textbf{98.12} & \textbf{80.17} & \textbf{84.66}       & \textbf{85.27}    & \textbf{43.34} & \textbf{60.55}      & \textbf{73.21}  \\
\hline
\multirow{2}{0.8cm}{Overcast} & HRNet      &      80.93                  &     68.84            &                 77.40 &      42.79        &  96.88   &  74.24     & 82.31            &         74.24 &   79.75    &     60.19       &    73.76     \\
& PointFlow  & \textbf{81.87} &  \textbf{70.28}               &                 \textbf{78.53} &    \textbf{46.22}          &   \textbf{97.00}  & \textbf{76.87}      &   \textbf{83.27}          &  \textbf{76.87}        &      \textbf{83.02} &    \textbf{65.11}        & \textbf{75.90}
\end{tabular}
\end{table*}

\section{Results} \label{re}

In this section, we assess the efficacy of selected multi-scale networks across both real and virtual datasets, focusing on segmentation accuracy as a pivotal metric of their performance. We analyze their precision in segmenting and classifying image elements and evaluates the influence of variable recording conditions, such as lighting and motion.

Furthermore, we explore the networks' ability in knowledge transfer, a crucial aspect determining their adaptability and generalizability over diverse datasets. This examination aims to compare their capacity for applying acquired insights across varying contexts, thereby highlighting their utility in a range of segmentation applications. Our goal is benchmarking the multi-scale networks for drone datasets.

\subsection{Setup}

When training the networks, we first take into account the problem of data imbalance, and compute the probability for each semantic class, as follows:

\begin{equation}
d=\frac{class\_pixels}{total\_pixels}*100\% 
\end{equation}

After obtaining the class distribution, we initialize the weights based on each class probability with the formula:

\begin{equation}
    weight_\mathrm{class} = \frac{1}{ln(c + probability_\mathrm{class})}
\end{equation} 

The data is divided into $60\%$ for training, $20\%$ for validation and $20\%$ for testing. We train the networks for 200 epochs on 4 NVIDIA Tesla V100 GPUs.  We report results for the metric IoU - Intersection over Union (also known as the Jaccard index), which measures the accuracy of an object detector on a particular dataset. It can be computed from the confusion matrix, which tells us the number of true positives -  the model correctly predicts the class, false positives - an outcome where the model incorrectly predicts the positive class, true negatives - the model correctly predicts the negative class, and false negatives -  model incorrectly predicts the negative class. The corresponding formula is: 
\begin{equation}
    IoU = TP/(TP+FP+FN)
\end{equation}

\subsection{Semantic Segmentation Results}

\begin{table}[!b]
\centering
\caption{Comparison of semantic segmentation performance, for HRNet and PointFlow network, on the sunny dataset, for each altitude and pitch degree.}\label{tab:sunny_results}
\resizebox{\columnwidth}{!}{
\begin{tabular}{l|cc|cc|cc}
Sequence          & \multicolumn{2}{c|}{\textbf{seq1}}         & \multicolumn{2}{c|}{\textbf{seq2}}           & \multicolumn{2}{c}{\textbf{seq3}}         \\
Altitude, pitch degree  & \multicolumn{2}{c|}{30m, 0°}     & \multicolumn{2}{c|}{30m, 60°}           & \multicolumn{2}{c}{30m, 90°}           \\
\hline
\diagbox[width=10em]{Class}{Network}             & HRNet   & PointFlow & HRNet    & PointFlow & HRNet    & PointFlow \\
\hline
Ground vegetation & 75.07   & 76.84     & 85.50    & 86.72     & 86.93    & 87.54     \\
Deciduous trees   & 68.66   & 70.28     & 72.91    & 74.18     & 74.01    & 76.12     \\
Coniferous trees  & 81.78   & 82.63     & 77.33    & 78.64     & 77.30    & 78.84     \\
Fallen trees      & 46.55   & 49.73     & 50.15    & 52.87     & 54.77    & 57.84     \\
Sky               & 95.07   & 95.21     & -        & -         & -        & -         \\
Rocks             & 73.33   & 75.28     & 79.28    & 81.78     & 80.75    & 83.16     \\
Dirt ground       & 65.65   & 67.03     & 83.92    & 85.07     & 87.13    & 89.75     \\
Building          & 70.52   & 74.84     & 88.39    & 90.73     & 91.68    & 93.72     \\
Fence             & 38.53   & 43.73     & 57.01    & 65.66     & 54.58    & 58.01     \\
Static car        & 39.42   & 43.81     & 65.39    & 70.14     & 73.47    & 77.54     \\
\hline
Overall           & 65.46   & 67.94     & 73.32    & 76.20     & 75.62    & 78.06     \\
\multicolumn{1}{l}{\textbf{}}\\
Sequence          & \multicolumn{2}{c|}{\textbf{seq4}}        & \multicolumn{2}{c|}{\textbf{seq5}} & \multicolumn{2}{c}{\textbf{seq6}}          \\
Altitude, pitch degree  & \multicolumn{2}{c|}{50m, 0°}           & \multicolumn{2}{c|}{50m, 60°}           & \multicolumn{2}{c}{50m, 90°}           \\
\hline
\diagbox[width=10em]{Class}{Network}             & HRNet   & PointFlow & HRNet    & PointFlow & HRNet    & PointFlow \\
\hline
Ground vegetation & 62.48   & 64.55     & 79.00    & 80.37     & 82.49    & 83.15     \\
Deciduous trees   & 56.12   & 59.12     & 70.19    & 72.03     & 72.66    & 73.44     \\
Coniferous trees  & 81.96   & 83.54     & 77.40    & 79.08     & 78.00    & 79.37     \\
Fallen trees      & 26.44   & 29.85     & 43.86    & 46.13     & 45.96    & 47.98     \\
Sky               & 97.73   & 97.67     & -        & -         & -        & -         \\
Rocks             & 74.87   & 76.18     & 77.01    & 78.21     & 78.62    & 81.34     \\
Dirt ground       & 52.07   & 54.86     & 80.19    & 81.47     & 86.12    & 87.72     \\
Building          & 64.18   & 69.88     & 83.55    & 85.54     & 89.18    & 92.74     \\
Fence             & 32.40   & 40.86     & 34.89    & 42.78     & 37.69    & 44.13     \\
Static car        & 28.63   & 35.58     & 59.86    & 61.37     & 64.61    & 67.86     \\
\hline
Overall           & 57.69   & 61.21     & 67.33    & 69.66     & 70.59    & 73.08     \\
\multicolumn{1}{l}{\textbf{}}\\
Sequence          & \multicolumn{2}{c|}{\textbf{seq7}}          & \multicolumn{2}{c|}{\textbf{seq8}} & \multicolumn{2}{c}{\textbf{seq9}}          \\
Altitude, pitch degree  & \multicolumn{2}{c|}{80m, 0°}           & \multicolumn{2}{c|}{80m, 60°}           & \multicolumn{2}{c}{80m, 90°}           \\
\hline
\diagbox[width=10em]{Class}{Network}             & HRNet   & PointFlow & HRNet    & PointFlow & HRNet    & PointFlow \\
\hline
Ground vegetation & 59.96   & 62.73     & 78.57    & 79.85     & 81.15    & 82.31     \\
Deciduous trees   & 48.34   & 51.35     & 64.72    & 67.16     & 69.24    & 70.88     \\
Coniferous trees  & 79.64   & 79.87     & 74.68    & 76.44     & 76.28    & 77.12     \\
Fallen trees      & 20.22   & 24.33     & 35.68    & 38.63     & 36.63    & 39.31     \\
Sky               & 98.64   & 98.45     & -        & -         & -        & -         \\
Rocks             & 75.50   & 77.32     & 75.90    & 79.56     & 73.59    & 76.17     \\
Dirt ground       & 50.90   & 53.94     & 77.86    & 79.67     & 84.09    & 85.62     \\
Building          & 58.71   & 63.87     & 81.73    & 84.11     & 86.78    & 88.46     \\
Fence             & 24.23   & 33.76     & 29.73    & 38.93     & 22.17    & 31.72     \\
Static car        & 26.10   & 30.74     & 54.68    & 62.31     & 61.36    & 66.24     \\
\hline
Overall           & 54.22   & 57.64     & 63.73    & 67.41     & 65.70    & 68.65    
\end{tabular}
}
\end{table}

\begin{table}[!b]
\centering
\caption{Comparison of semantic segmentation performance, for HRNet and PointFlow network, on the overcast dataset, for each altitude and pitch degree.}\label{tab:overcast_results}
\resizebox{\columnwidth}{!}{
\begin{tabular}{l|cc|cc|cc}
Sequence          & \multicolumn{2}{c|}{\textbf{seq10}}           & \multicolumn{2}{c|}{\textbf{seq11}}           & \multicolumn{2}{c}{\textbf{seq12}}           \\
Altitude, pitch degree & \multicolumn{2}{c|}{30m, 0°}     & \multicolumn{2}{c|}{30m, 60°}           & \multicolumn{2}{c}{30m, 90°}       \\
\hline
\diagbox[width=10em]{Class}{Network}             & HRNet   & PointFlow & HRNet    & PointFlow & HRNet    & PointFlow \\
\hline
Ground vegetation & 77.73   & 78.88     & 85.89    & 86.50     & 86.81    & 87.43     \\
Deciduous trees   & 70.99   & 72.52     & 71.59    & 72.66     & 73.94    & 74.85     \\
Coniferous trees  & 80.48   & 81.38     & 74.80    & 75.98     & 75.44    & 76.63     \\
Fallen trees      & 42.02   & 44.52     & 50.17    & 53.13     & 50.25    & 55.18     \\
Sky               & 94.31   & 94.46     & -        & -         & -        & -         \\
Rocks             & 71.19   & 73.35     & 72.70    & 75.38     & 79.75    & 82.16     \\
Dirt ground       & 64.69   & 66.03     & 84.02    & 84.62     & 86.28    & 86.94     \\
Building          & 65.32   & 69.25     & 84.17    & 86.45     & 87.34    & 90.27     \\
Fence             & 36.35   & 43.77     & 39.87    & 49.10     & 42.80    & 53.89     \\
Static car        & 34.85   & 41.64     & 65.54    & 69.89     & 70.92    & 74.83     \\
\hline
Overall           & 63.79   & 66.58     & 69.86    & 72.63     & 72.61    & 75.80     \\
\multicolumn{1}{l}{\textbf{}}\\
Sequence          & \multicolumn{2}{c|}{\textbf{seq13}}              & \multicolumn{2}{c|}{\textbf{seq14}}           & \multicolumn{2}{c}{\textbf{seq15}}           \\
Altitude, pitch degree   & \multicolumn{2}{c|}{50m, 0°}           & \multicolumn{2}{c|}{50m, 60°}           & \multicolumn{2}{c}{50m, 90°}   \\
\hline
\diagbox[width=10em]{Class}{Network}             & HRNet   & PointFlow & HRNet    & PointFlow & HRNet    & PointFlow \\
\hline
Ground vegetation & 58.58   & 60.78     & 79.35    & 80.31     & 82.19    & 82.90     \\
Deciduous trees   & 56.57   & 58.94     & 70.04    & 71.16     & 72.87    & 73.61     \\
Coniferous trees  & 82.26   & 83.16     & 76.15    & 77.27     & 77.16    & 78.15     \\
Fallen trees      & 24.05   & 27.29     & 41.15    & 45.15     & 45.52    & 48.98     \\
Sky               & 97.23   & 97.33     & -        & -         & -        & -         \\
Rocks             & 72.94   & 75.28     & 74.31    & 76.94     & 74.97    & 77.74     \\
Dirt ground       & 51.97   & 54.62     & 81.22    & 82.23     & 85.94    & 86.68     \\
Building          & 60.40   & 65.92     & 82.44    & 85.26     & 87.49    & 89.61     \\
Fence             & 27.49   & 36.95     & 29.33    & 38.45     & 26.19    & 36.44     \\
Static car        & 28.48   & 34.39     & 56.22    & 61.55     & 70.59    & 74.89     \\
\hline
Overall           & 56.00   & 59.47     & 65.58    & 68.70     & 69.21    & 72.11     \\
\multicolumn{1}{l}{\textbf{}}\\
Sequence          & \multicolumn{2}{c|}{\textbf{seq16}}         & \multicolumn{2}{c|}{\textbf{seq17}}           & \multicolumn{2}{c}{\textbf{seq18}}           \\
Altitude, pitch degree  & \multicolumn{2}{c|}{80m, 0°}           & \multicolumn{2}{c|}{80m, 60°}           & \multicolumn{2}{c}{80m, 90°}           \\
\hline
\diagbox[width=10em]{Class}{Network}             & HRNet   & PointFlow & HRNet    & PointFlow & HRNet    & PointFlow \\
\hline
Ground vegetation & 54.05   & 56.94     & 76.43    & 77.82     & 80.41    & 81.35     \\
Deciduous trees   & 49.17   & 52.56     & 64.70    & 66.77     & 69.42    & 70.72     \\
Coniferous trees  & 79.05   & 80.52     & 73.78    & 75.21     & 75.09    & 76.20     \\
Fallen trees      & 22.97   & 25.01     & 34.37    & 37.33     & 35.99    & 39.24     \\
Sky               & 98.30   & 98.45     & -        & -         & -        & -         \\
Rocks             & 74.15   & 76.68     & 74.70    & 77.52     & 71.01    & 74.20     \\
Dirt ground       & 51.60   & 54.01     & 79.88    & 81.35     & 85.84    & 86.63     \\
Building          & 54.95   & 61.13     & 78.40    & 81.89     & 83.23    & 86.32     \\
Fence             & 19.99   & 29.43     & 21.37    & 31.56     & 16.97    & 24.34     \\
Static car        & 21.41   & 27.68     & 52.00    & 59.89     & 61.15    & 67.07     \\
\hline
Overall           & 52.56   & 56.24     & 61.74    & 65.48     & 64.35    & 67.34    
\end{tabular}
}
\end{table}

The training results for the real dataset can be seen in Table \ref{tab:wilduav_results}, from which we can notice that the best performance is obtained for the classes with the largest percentage of data, such as ground vegetation and trees. The worst performing classifications, for both networks, are for classes people and moving car. We notice that PointFlow network performs better overall by $+2.40$, since it has the advantage of computing boundary areas. This is most evident for classes like cars, where it obtains an improvement of $+5.23$, people, dirt ground and empty performing better by $+3$, while the rest of the classes are also seeing corrections.

In the case of the synthetic dataset, Table \ref{tab:fi_sunny} presents the segmentation results on the sunny and overcast lighting conditions. The highest accuracy is obtained for classes sky, building, ground vegetation and trees, while the class fence posed difficulties to both frameworks. Point Flow network performed better by $+2.37$ than HRNet, with the most significant differences being noticed for labels fence, rocks, fallen trees, ground vegetation, and static car. For the overcast scenario, PointFlow network obtains an improvement of $+2.14$ overall, with the most noticeable differences for classes static car, fallen tree and fence. This improvement is gained from the edge information, that allows to correctly identify object boundaries. Combining the sunny and overcast test results, we observe that PointFlow network performs better than HRNet, on average with $3.12\%$.

When studying the impact of transfer learning, we notice from the second half of Table \ref{tab:wilduav_results} that first training on a virtual dataset and then refining the network on the real drone data performs better for both networks. This is useful if a dataset lacks certain classes, or some labels are insufficiently represented.

\subsection{Ablation Study}

Table \ref{tab:ablation_results} presents the two networks in parallel, comparing the number of parameters, the total memory used and the runtime in milliseconds. We notice that HRNet is more lightweight, with only 1.5 million parameters, while Point Flow net uses 33 million. This also affects the size of the model, which is considerably bigger in the case of the second architecture. The inference time needed to semantically segment a $512\times512$ image is 3 times higher for PointFlow network.

To study the impact of altitude and camera pitch angle, we created Tables \ref{tab:sunny_results} and \ref{tab:overcast_results}. We notice that the networks perform better when the camera angle is oriented towards the ground, at -90°, which can be inferred from the similarity of object shape and size. The advantage of a bird-eye view comes from the lack of perspective transformation on the objects, thus the segmentation results are more accurate. Similarly, the networks' performance decreases with the increase of recording altitude.

Comparing the results from Table \ref{tab:fi_sunny} with the ones from Table \ref{tab:sunny_results}, respectively Table \ref{tab:overcast_results}, we observe that when we work with a dataset that contains recordings with varying altitudes and camera angles, the best strategy to employ is to train on all data at once, rather than independently. This allows a holistic view of the area, and the network can learn the representation of the same objects, at different dimensions and angles of view.

\begin{table}[!t]
\centering
\caption{Comparison of networks in terms of number of parameters used, size computed in GFlops, and inference time for an image of 512 x 512 pixels.}\label{tab:ablation_results}
\begin{tabular}{l|ccc}
Network       & \# params (M) & Size (GFlops) & Time (ms) \\
\hline
HRNet         & 1.5                    & 3.8           & 13                  \\
PointFlow net & 33                     & 85.8          & 44                 
\end{tabular}
\end{table}

\begin{figure}[t]
	\includegraphics[width=\columnwidth]{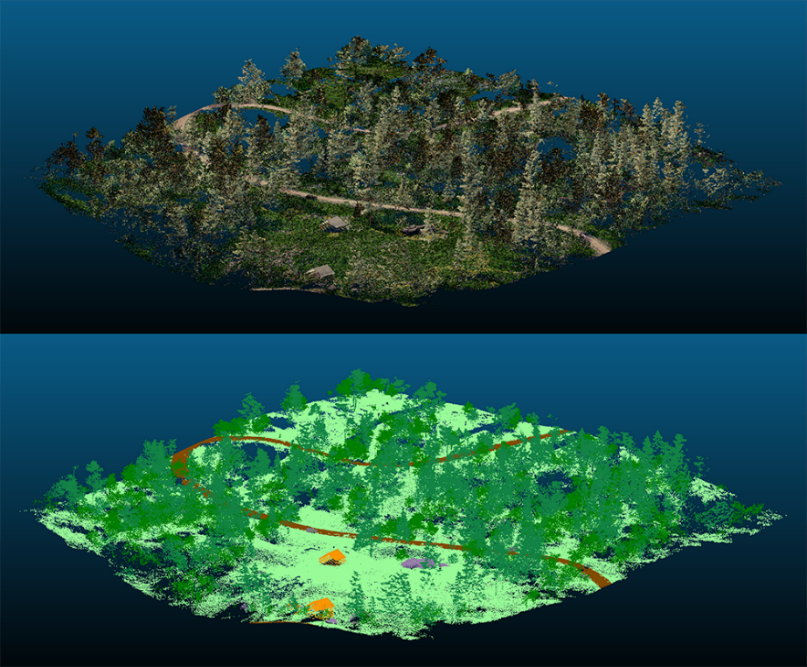}
	\centering
	\caption{Point clouds recorded in the virtual environment, with color (top) and semantic (bottom) information, side view.}
	\label{fig:image9}
			\vspace*{-3mm}
\end{figure}

\begin{figure}[t]
	\includegraphics[width=0.9\columnwidth]{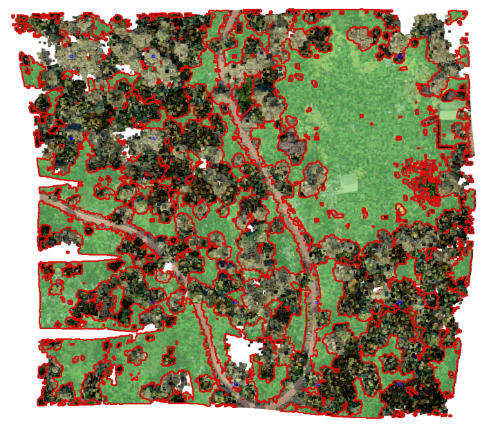}
	\centering
	\caption{Extraction of regions without trees in the virtual map. The information on the degree of deforestation in relation to the analyzed area tells us that 53.70\% is deforested area, the percentage of healthy trees is 39.73\% and 0.48\% are fallen trees.}
	\label{fig:image11}
			\vspace*{-5mm}
\end{figure}

\section{Deforestation Assessment} \label{da}

Based on the virtual environment in the simulator, we developed a methodology to describe the degree of deforestation of an area. The first step is to map the forest, which is done by recording color and semantic images from a drone flying over the area. We use sequence 6 from the Forest Inspection Dataset, which was recorded from the drone at an altitude of 50 meters with the camera facing the ground. 

The second step is the construction of a 3D point cloud with color and semantic segmentation information, for which we used the library presented in \cite{alv2021}, called Open3D. To create a point cloud we use the RGBD image and the intrinsic and extrinsic camera matrices. The first matrix is used to generate the point cloud from the color and depth information in the camera frame, while the second matrix is needed to stitch together multiple point clouds. Each pixel from the RGBD image is represented by a color (either from the image, either from the semantic segmentation) and a depth value, while the intrinsic matrix is defined as:

\begin{equation}
K=\begin{pmatrix}
  f_x & 0 & c_x \\
  0 & f_y & c_y \\
  0 & 0 & 1   
	\end{pmatrix}
\end{equation}
\begin{equation}
fov_{rad}=\frac{fov*\pi}{180}
\end{equation}
\begin{equation}
f_x=f_y=\frac{w/2.0}{\tan(fov_{rad}/2.0)}
\end{equation}
\begin{equation}
c_x=w/2-0.5, c_y=h/2-0.5
\end{equation}
where $w$ is the image width, $h$ is the image height, $fov$ is the field of view (90°, in our case).

\begin{figure}[!b]
	\includegraphics[width=0.8\columnwidth]{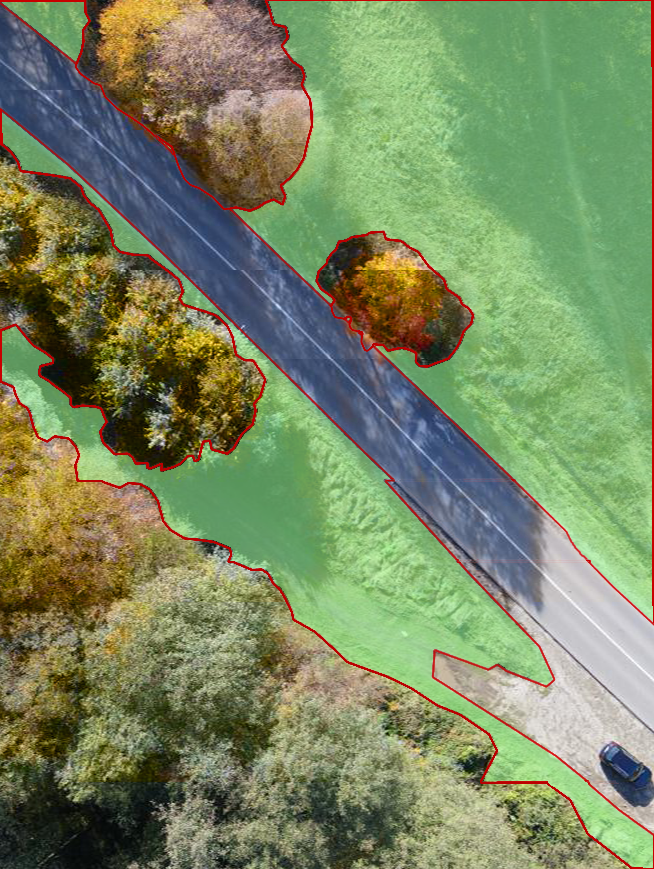}
	\centering
	\caption{Extraction of regions without trees in the real environment. 38.37\% is ground vegetation, while 41.62\% is covered by healthy trees.}
	\label{fig:dde}
\end{figure}

To compute the extrinsic matrix, we determine the translation and rotation matrices, from the positioning information provided in the \textit{airsim\_rec.txt} file. For each image, we know the position of the drone in $x$, $y$ and $z$ coordinates, and the rotation quaternion given by $q_w$, $q_x$, $q_y$ and $q_z$. The translation matrix is given by:
\begin{equation}
T=\begin{pmatrix}
  -y & 0 & 0 & 0 \\
  0 & -z & 0 & 0\\
  0 & 0 & -x  & 0 \\
  0 & 0 & 0 & 1
	\end{pmatrix}
\end{equation}

While the rotation matrix can be computed as:
\begin{equation}
R=\begin{pmatrix}
  2(q_0^2+q_1^2)-1 & 2(q_1q_2-q_0q_3) & 2(q_1q_3+q_0q_2) & 0 \\
  2(q_1q_2+q_0q_3) & 2(q_0^2+q_2^2)-1 & 2(q_2q_3-q_0q_1) & 0\\
  2(q_1q_3-q_0q_2) & 2(q_2q_3+q_0q_1) & 2(q_0^2+q_3^2)-1 & 0 \\
  0 & 0 & 0 & 1
	\end{pmatrix}
\end{equation}

Thus the extrinsic matrix is computed as:
\begin{equation}
F=R^\intercal T C
\end{equation}

Where $C$ is a matrix used to transform the point cloud into a coordinate system for point cloud reconstruction: 
\begin{equation}
C=\begin{pmatrix}
  1 & 0 & 0 & 0 \\
  0 & 0 & -1 & 0\\
  0 & 1 & 0  & 0 \\
  0 & 0 & 0 & 1
	\end{pmatrix}
\end{equation}

From the Open3D library, to compute a single point cloud, we use the function \textit{create\_point\_cloud\_from\_rgbd\_image} from the package \textit{geometry}, which takes as input the RGBD image, the intrinsic matrix $K$, and the extrinsic matrix $F$. We can then obtain the full mesh by adding the generated point clouds since all of them are represented in the same world coordinates. As seen in Figure \ref{fig:image9}, we obtain the representation of the natural environment, together with the semantic information. 

Using the generated semantic point cloud, we create a snapshot from bird's-eye view, in which we have the representation of the studied area in terms of class labels. The accuracy of the depth information ensured that even if there was overlap between consecutive recordings, there is consistency in the final result. Studying the full area is more representative than assessing each image independently, giving us a holistic view. 

The next step is to apply a morphological dilation operation by which we obtain a denser map of semantic information, based on which we extracted the contours of the deforested areas. By spatially merging the recordings, we obtain a comprehensive map of the desired location, from which we can extract information like the percentage of healthy trees, fallen trunks and deforested areas, as seen in Figure \ref{fig:image11}. This methodology can be applied to the same area at different time intervals, to obtain the evolution of the forest. This method can be successfully applied to real-world recordings, as showcased in Figure \ref{fig:dde}.

\section{Conclusion} \label{conclusion}

In this paper, we introduce the novel Forest Inspection dataset which combines real and virtual recordings from natural environments, using drones. The recordings contain color images, enriched with depth and positioning information, to facilitate a comprehensive suite of tasks encompassing semantic segmentation, depth perception analysis, deforestation monitoring, and the facilitation of autonomous navigation systems.  Through our exploration of state-of-the-art methodologies within the field of multi-scale networks, we have found that a hierarchical, pyramidal framework significantly outperforms traditional approaches, especially given the dataset's diverse array of altitudes and camera perspectives.

Additionally, by leveraging the depth data included within our dataset, we have successfully generated 3D point cloud reconstructions. This enables a higher level of analysis, allowing for the precise quantification of arboreal health, including the accurate determination of both healthy and fallen trees. Such insights are valuable for environmental monitoring and the proactive management of natural resources, underscoring our dataset's utility and relevance in the context of ecological studies and sustainability efforts.

In the future, we want to enhance the semantic segmentation network with depth information, to increase the accuracy on all altitudes. We also want to obtain a voxel representation of the environment, and to apply reinforcement learning that would allow a drone to fly over a forest at different time intervals to automatically monitor its evolution.

\section*{Acknowledgments}
This work was supported by the “SEPCA-Integrated Semantic Visual Perception and Control for Autonomous Systems” grant funded by the Romanian Ministry of Education and Research, code PN-III-P4-ID-PCCF-2016-0180.

\bibliographystyle{IEEEtran}  
\bibliography{main.bib}

\begin{IEEEbiography}[{\includegraphics[width=1in,height=1.25in,clip,keepaspectratio]{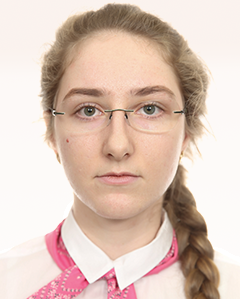}}]{Bianca-Cerasela-Zelia Blaga}
received the M.S. degree from the Technical University of Cluj-Napoca (TUCN), Cluj-Napoca, Romania, in 2019 in the field of Artificial Intelligence and Vision. She is pursuing the Ph.D. degree with the Faculty of Automation and Computer Science, TUCN, in the domain of visual perception and scene understanding for autonomous systems. Her research interests include image processing, artificial intelligence, scene understanding, and autonomous vehicles.

\end{IEEEbiography}

\begin{IEEEbiography}[{\includegraphics[width=1in,height=1.25in,clip,keepaspectratio]{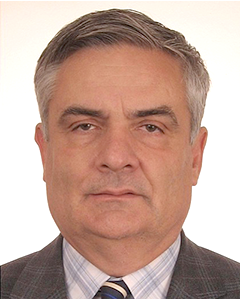}}]{Sergiu Nedevschi}
received the M.S. and Ph.D. degrees in electrical engineering from the Technical University of Cluj-Napoca (TUCN), Cluj-Napoca, Romania, in 1975 and 1993, respectively. From 1976 to 1983, he was a Researcher with the Research Institute for Computer Technologies, Cluj-Napoca. In 1983, he joined TUCN, where he was appointed as a Full Professor of computer science in 1998 and founded and since then he has been leading the Image Processing and Pattern Recognition Research Center. From 2000 to 2004, he was the Head of the Computer Science Department, from 2004 to 2012, he was the Dean of the Faculty of Automation and Computer Science, and from 2012 to 2020, he was the Vice-Rector with the Scientific Research of TUCN. He was involved in more than 80 research projects, being the coordinator of 62 of them. His industrial cooperation with important automotive players, such as Volkswagen AG, Robert Bosch GmbH, and SICK AG all from Germany, and research institutes, such as VTT from Finland and INRIA from France, was achieved through funded research projects. He has published more than 400 scientific articles and has edited over 20 volumes, including books and conference proceedings. His research interests include image processing, pattern recognition, computer vision, machine learning, intelligence, and autonomous vehicles..
\end{IEEEbiography}

\vfill

\end{document}